\documentclass[]{spie}  

\usepackage{amsmath,amsfonts,amssymb}
\usepackage{tensor}
\usepackage{graphicx}
\usepackage{units}
\usepackage[colorlinks=true, allcolors=blue]{hyperref}
\usepackage{subfig}

\usepackage{indentfirst}

\usepackage{mathptmx}
\usepackage[T1]{fontenc}

\usepackage[labelsep=period]{caption}

\title{Spatio-thermal depth correction of RGB-D sensors based on Gaussian Processes in real-time} %

\author[a]{Christoph Heindl}
\author[a]{Thomas P\"{o}nitz}
\author[a]{Gernot St\"{u}bl}
\author[a]{Andreas Pichler}
\author[b]{Josef Scharinger}
\affil[a]{Profactor GmbH, Im Stadtgut A2, 4407 Steyr-Gleink, Austria}
\affil[b]{JKU Department of Computational Perception, Altenbergerstr. 69, 4040 Linz, Austria}

\newcommand{\etal}{et al.}
\newcommand{\degree}{$^{\circ}$}
\newcommand{\Depth}[3]{\tensor*[^{#2}_{#3}]{\mathcal{D}}{_{\mathrm{#1}}}}

\begin{document} 

\maketitle

\begin{abstract}
Commodity RGB-D sensors capture color images along with dense pixel-wise depth information in real-time. Typical RGB-D sensors are provided with a factory calibration and exhibit erratic depth readings due to coarse calibration values, ageing and thermal influence effects. This limits their applicability in computer vision and robotics. We propose a novel method to accurately calibrate depth considering spatial and thermal influences jointly. Our work is based on Gaussian Process Regression in a four dimensional Cartesian and thermal domain. We propose to leverage modern GPUs for dense depth map correction in real-time. For reproducibility we make our dataset and source code publicly available.
\end{abstract}


\section{INTRODUCTION}
\label{sec:intro} 

Since their introduction in 2010 active structured-light RGB-D cameras have been widely applied to computer vision and robotic applications \cite{zhang2012microsoft}. Newcombe \etal\cite{newcombe2011kinectfusion} successfully employed RGB-D cameras in real-time 3D surface reconstructions. Later this work was extended by Heindl \etal\cite{heindl2016capturing} to generate photorealistic reconstructions by aligning depth and color streams. Combining features from depth and color images robustified object recognition\cite{lai2011sparse} and led to a transition from stereo vision to RGB-D cameras as the main imaging device in indoor robotics\cite{el2012study}.

RGB-D cameras provide simultaneous streams of color and dense depth information. Color images are typically provided by standard RGB cameras, while depth readings are generated by a separate time-of-flight or active structured infrared light sensor. Both devices are usually assembled together in a compact housing. A default factory calibration encompassing intrinsic and extrinsic camera parameters is stored within the device memory. The quality of calibration is often impractical for computer vision applications with high accuracy demands for the following reasons: the assumed pinhole camera model together with radial and tangential distortion coefficients does not capture the true underlying nonlinear deformations found in depth maps well. Additionally, strong thermal dependencies on depth readings undermine the effectiveness of a single static mono/stereo calibration.

In this paper we propose a novel calibration method that corrects depth readings based on a nonlinear model that spans the spatial and thermal domain. We frame depth correction as a probabilistic regression problem and show that Gaussian Process Regression (GPR) is well suited for capturing the underlying systematic errors. For dense real-time correction we propose to leverage the fact that GPR prediction is mostly linear algebra and is therefore well suited to be parallelized on modern GPU architectures. Python code alongside with a captured dataset is made publicly available\footnote{\url{https://github.com/cheind/rgbd-correction}}.

\section{RELATED WORK}
\label{sec:related}
The availability of modern commodity RGB-D sensors raised the interest in methods for accurate calibration thereof. Typical approaches \cite{konolige2012technical, burrus2011kinect} use a calibration target together with intrinsic and extrinsic parameter estimation methods, such as the one proposed by Zhang \cite{zhang2000flexible}. Since one usually has no access to the internals of the depth generation process, these attempts are insufficient for accurate depth correction, because they merely correct for lens effects. In contrast, the work of Smisek et al. \cite{smisek20133d} assumes planarity in the observed data. They propose to correct depth by computing a pixel-wise mean residual depth image over all calibration poses. Zhang et al. \cite{zhang2014calibration} exploit co-planarity in the structure of the calibration target and propose to model the observed depth as a linear function of the true depth. In the work of Canessa et al. \cite{canessa2014calibrated} a second order polynomial per pixel is proposed to compensate for depth artefacts. The closest work compared to ours, in terms of applied methods, is by Amamra et. al \cite{amamra2014rgbd}, who use a Gaussian Process to predict absolute depth from spatial locations. More recently online depth calibration methods try to compensate depth errors using a visual SLAM system\cite{quenzel2017}.

Several research groups documented a strong thermal influence on the depth generation process of RGB-D sensors. Mankoff et al. \cite{mankoff2013kinect} note a severe depth shift due to internal or external thermal changes. Fiedler et al. \cite{fiedler2013impact} document a nonlinear distortion effects during thermal changes. They also propose practical rules of thumb on reducing accuracy errors caused by thermal conditions. 

To our knowledge we are the first to propose a practical, real-time, pixel-wise depth correction method for RGB-D cameras that considers spatial and thermal aspects jointly.

\section{DEPTH CORRECTION}
\label{sec:depthcorrection}
Throughout this work we use lower-case non-bold characters $x$ to denote scalars, bold-faced lower-case characters $\mathbf{x}$ represent column-vectors and upper-case bold characters $\mathbf{A}$ for matrices. $\mathbf{x}_i$ denotes the $i$-th element of $x$, $\mathbf{A}_{ij}$ the $i$-th row and $j$-th column of $\mathbf{A}$. Functions will be written in calligraphic letters $\mathcal{F}$.

\subsection{Dense depth maps}
\label{sec:densedepthmaps}
RGB-D cameras provide depth readings in dense image form $\mathcal{D} \colon \mathbb{Z}^2 \rightarrow \mathbb{R}$. Depth values can be reprojected into three-dimensional Cartesian space through a function $\mathcal{X}\colon \mathbb{Z}^2 \times \mathbb{R} \rightarrow \mathbb{R}^{3 \times 1}$ given by
\begin{equation}
\label{eq:reproject}
\mathcal{X}(i,j,d) = d\mathbf{K}^{-1}\mathbf{h}
\end{equation}
where $\mathbf{K} \in \mathbb{R}^{3 \times 3}$ is the associated camera intrinsic matrix and $\mathbf{h}=[i, j, 1]^T$ is a point in homogeneous image space. Typically one is interested in alignment of depth and RGB images. This can be accomplished by transforming the entire depth map into the RGB camera space as follows
\begin{equation}
\begin{aligned}
\label{eq:mapdepth}
\mathbf{x} &= \mathcal{X}(i,j,\mathcal{D}(i,j))\\
\mathbf{x}' &= \mathbf{K}'(\mathbf{R}\mathbf{x} + \mathbf{t})\\
\mathcal{D}'(\mathbf{x}'_0,\mathbf{x}'_1) &= \mathbf{x}'_2
\end{aligned}
\end{equation}
where $\mathbf{R} \in \mathbb{R}^{3\times3}$ and $\mathbf{t} \in \mathbb{R}^{3\times1}$ are parameters of the rigid extrinsic transformation matrix, $\mathbf{K}' \in \mathbb{R}^{3\times3}$ is the intrinsic camera matrix of the target sensor. Extrinsic camera parameters are either obtained by stereo camera calibration or are directly provided by the device itself. We do not expect intrinsic nor extrinsic matrices to be very accurate for our approach to work.

\subsection{Gaussian Process Regression}
\label{sec:gpr}
A Gaussian Process (GP)\cite{rasmussen2006gaussian,murphy2012machine} is a statistical model used in supervised machine learning for regression and classification tasks. In regression, one searches for a unknown function $\mathcal{F}(\mathbf{X}) = \mathbf{y} + \epsilon $ that best approximates tuples of feature vectors $\mathbf{X} \in \mathbb{R}^{N \times d}$, and target scalar observations $\mathbf{y} \in \mathbb{R}^{N \times 1}$. The term $\epsilon \sim \mathcal{N}(0, \sigma_y^2)$ is the usual i.i.d. Gaussian observation noise. In a probabilistic view, the expectation of the posterior predictive distribution is used to infer unknown values $\mathbf{y}_{\star}$ from feature vectors $\mathbf{X_{\star}}$. The posterior predictive takes on the following general form
\begin{equation}
	\label{eq:generalposteriorpredictive}
	p(\mathbf{y}_{\star}\vert \mathbf{X_{\star}}, \mathbf{X}, \mathbf{y}) = \int p(\mathbf{y_{\star}}\vert \mathcal{F},\mathbf{X_{\star}}) p(\mathcal{F}\vert\mathbf{X}, \mathbf{y})d\mathcal{F}
\end{equation}
where $\mathbf{y}_{\star} \in \mathbb{R}^{M \times 1}$ and $\mathbf{X_{\star}} \in \mathbb{R}^{M \times d}$. Typically, integrating over all possible functions is intractable. Hence, Gaussian Processes make the assumption that training and predictions are jointly Gaussian
\begin{equation}
	\label{eq:jointgaussian}
	\begin{bmatrix}
		\mathbf{y}\\\mathbf{y_{\star}}
	\end{bmatrix}
	\sim 
	\mathcal{N}\left(
		\begin{bmatrix}
			\mathcal{M}(\mathbf{X})\\
			\mathcal{M}(\mathbf{X_{\star}})\\
		\end{bmatrix},
		\begin{bmatrix}
			\Sigma(\mathbf{X}, \mathbf{X}) & \Sigma(\mathbf{X}, \mathbf{X_{\star}}) \\ 
			\Sigma(\mathbf{X}, \mathbf{X_{\star}})^T & \Sigma(\mathbf{X_{\star}}, \mathbf{X_{\star}}) 
		\end{bmatrix}
	\right)
\end{equation}
where $\mathcal{M} \colon \mathbb{R}^{K \times d} \rightarrow \mathbb{R}^{K \times 1}$ is the mean function and  $\Sigma \colon \mathbb{R}^{K \times d} \times \mathbb{R}^{L \times d} \rightarrow \mathbb{R}^{K \times L}$ is a positive-definite function. Then, from Gaussian conditioning rules \cite{roweis1999gaussian} one gets the posterior predictive in closed form by conditioning on $\mathbf{X_{\star}}$, $\mathbf{X}$ and $\mathbf{y}$
\begin{equation}
	\label{eq:posteriorpredictive}
	\begin{aligned}
	p(\mathbf{y_{\star}}\vert \mathbf{X_{\star}}, \mathbf{X}, \mathbf{y}) 
	&\sim 
	\mathcal{N}(\mathbf{u_{\star}}, \mathbf{K_{\star}}), \\
	\mathbf{u_{\star}} &= \mathcal{M}(\mathbf{X_{\star}}) + \Sigma(\mathbf{X}, \mathbf{X_{\star}})^T\Sigma(\mathbf{X}, \mathbf{X})^{-1}(\mathbf{y} - \mathcal{M}(\mathbf{X})), \\
	\mathbf{K_{\star}} &= \Sigma(\mathbf{X_{\star}}, \mathbf{X_{\star}}) - \Sigma(\mathbf{X}, \mathbf{X_{\star}})^T\Sigma(\mathbf{X}, \mathbf{X})^{-1}\Sigma(\mathbf{X}, \mathbf{X_{\star}}).
	\end{aligned}
\end{equation} 
The covariance function $\Sigma$ encodes a distance or similarity measure between input samples.  For two samples $i$, $j$, $\Sigma$ is given by $\Sigma_{ij} = \sigma_s^2\mathcal{K}(\mathbf{x_i}, \mathbf{x_j}) + \delta_{ij}\sigma_y^2$, with $\mathcal{K}$ being a positive definite kernel function. Multiple kernel functions are available\cite{duvenaud2014automatic}. Choosing one is highly problem specific. In this work we focus on the smooth radial basis function (RBF) a.k.a. the squared exponential kernel which, in the multidimensional input case, is given by 
\begin{equation}
\label{eq:covariancekernel}
\Sigma(\mathbf{x_i},\mathbf{x_j}) = \sigma_s^2 e^{-0.5(\mathbf{x}_i-\mathbf{x}_j)^T\mathbf{W}(\mathbf{x}_i-\mathbf{x}_j)} + \delta_{ij}\sigma_y^2
\end{equation}
where $\sigma_s^2$ is the signal variance regularizing the allowed deviation from the mean,  $\mathbf{W} \in R^{d \times d}$ is an input domain coordinate scaling matrix, and $\sigma_y^2$ is the expected noise level. We restrict our considerations to diagonal versions of $\mathbf{W}$. The effects of varying these parameters are illustrated in \cite{murphy2012machine}.

\subsection{Approach for Spatio-Thermal Correction }

\label{sec:approach}
Our approach for correcting depth maps is based on estimating a per-pixel depth correction offset. For a given depth map $\mathcal{D}$ and temperature reading $t$ we compute a corrected depth map $\mathcal{D_{\star}}$ by 
\begin{equation}
	\label{eq:basiccorrection}
	\mathcal{D_{\star}}(i,j) = \mathcal{D}(i,j) + \Delta_{ij}
\end{equation}
where $\Delta_{ij}$ is given by a Gaussian Process Regression $\Delta_{ij} = \mathcal{G}(\mathbf{x}_{ij})$. The input $\mathbf{x}$ to the Gaussian Process $\mathcal{G}$ is given by reprojecting the pixel coordinate to Cartesian space using pixel-wise (potentially distorted) depth information and stacking the result with $t$ to create a four-dimensional input feature vector
\begin{equation}
	\label{eq:inputx}
	\mathbf{x}_{ij} = 
	\begin{bmatrix}
		x\\y\\z\\t
	\end{bmatrix} = 
	\begin{bmatrix}
		\mathcal{D}(i,j)\mathbf{K}^{-1}[i,j,1]^T \\
		t
	\end{bmatrix}.
\end{equation}

The corresponding regression target at training time $y_{ij}$ is given by the pixel-wise difference between a ground-truth depth map $\mathcal{D}_{\mathrm{gt}}(i,j)$ and observed depth map $\mathcal{D}_{\mathrm{obs}}(i,j)$, where
\begin{equation}
	\label{eq:inputy}
	y_{ij} = \mathcal{D}_{\mathrm{gt}}(i,j) - \mathcal{D}_{\mathrm{obs}}(i,j).
\end{equation}

Regressing relative correction offsets leads to predictions that vary smoothly accross the input domain. This justifies our usage of the squared exponential kernel given in Equation \ref{eq:covariancekernel}, which has a strong prior on smooth functions. Note, this is in contrast to the approach taken in Amamra et al.\cite{amamra2014rgbd}, who regress absolute depth using a similar kernel. However, in our experience real world depth maps rarely exhibit smooth behaviour due to strong depth discontinuities along object boundaries.

At training time we sample feature vectors and regression values in the spatio-thermal domain using a regular spaced grid. At test time we compute for every four dimensional query point $\mathbf{x}_{ij}$ the optimal prediction $\Delta_{ij} = \mathbf{u}_{\star\mathrm{ij}}$ and confidence $\mathbf{K_{\star\mathrm{ij}}}$ according to Equation~\ref{eq:posteriorpredictive}. In turn $\Delta_{ij}$ is used to construct the corrected depth map $\mathcal{D_{\star}}(i,j)$ according to Equation~\ref{eq:basiccorrection}.

\pagebreak
\section{EVALUATION}
This section covers the experiments which have been performed to prove the applicability of the approach.

\subsection{Hardware Setup}
\label{sec:hardware}
\begin{figure} [htp]
	\begin{center}
		\begin{tabular}{c} 
			\includegraphics[width=0.45\textwidth ]{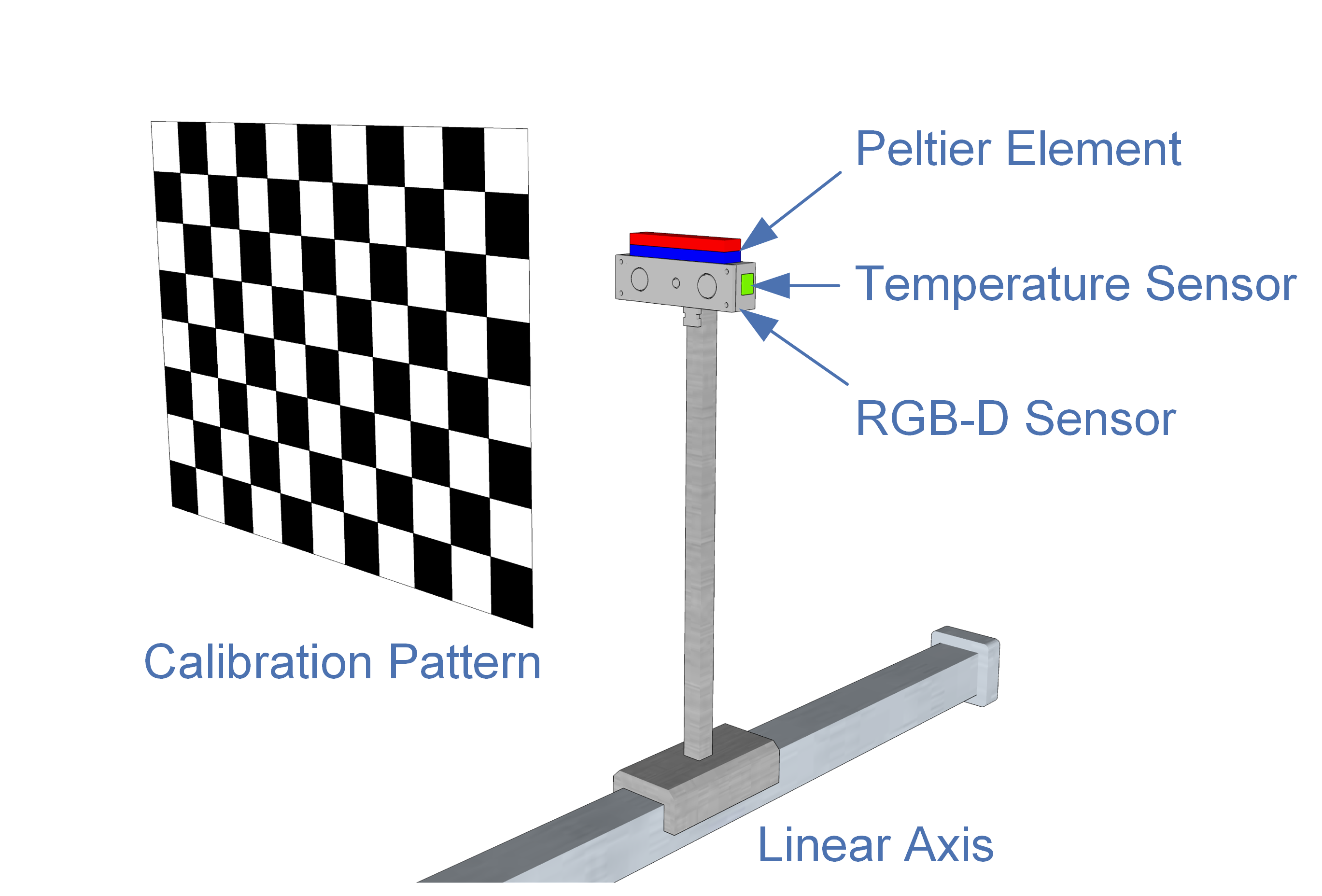}
		\end{tabular}
	\end{center}
	\caption[Schematic overview of hardware setup.] 
	{ \label{fig:hwSetup} The RGB-D camera is mounted on a linear axis and observes a calibration pattern at different distances. A Peltier element for heating and cooling is thermally bonded to the heat sink of the camera as well as the temperature sensor of the control loop.}
\end{figure}

Figure~\ref{fig:hwSetup} is a schematic overview of the hardware set-up used for data acquisition. As RGB-D sensor an Orbbec Astra Mini S\footnote{\url{https://orbbec3d.com/astra-mini/}} is used, since it is a typical commodity sensor without special housing. The sensor looks towards a chessboard calibration pattern and is mounted on an electronic linear axis to adjust the distance.
Directly connected to the heat sink of the sensor is a Peltier element which is used for heating and cooling. Due to the fact that all temperature-critical elements of the sensor are connected to the heat pipe, it is assumed that thermostabilization of the heat pipe directly controls the temperature of the main camera components, e.g image sensors and projector. 

The sensor of the temperature control loop is directly thermally coupled to the heat pipe, which is of special importance. In a future application, no control loop is needed and the sensor is the only additional hardware which is required to apply the proposed correction approach. Since the housing of an Orbbec Astra Mini S serves as heat pipe, there is no need to modify any sensor hardware in an application.

The evaluation itself is run on a workstation with an Intel i7-7700 @ \unit[3.60]{GHz}, \unit[16]{GB} RAM, and a NVIDIA GeForce GTX1080 graphics card with \unit[6]{GB} memory.

\subsection{Spatio-thermal RGB-D dataset}
\label{sec:dataset}
Using the hardware setup described above we captured the following data: for each temperature level, ranging from 10 to \unit[35]{\degree}C in steps \unit[1]{\degree}C, we moved the axis to six different positions spaced \unit[10]{cm} apart. The covered  distance between sensor and calibration target ranges from 0.4 to 1 meter, which matches the operating specifications of the sensor. At each calibration position we captured 50 RGB and 50 depth frames. Depth data was captured aligned with color data as described in Section~\ref{sec:densedepthmaps}. From each color frame an artificial dense depth map was computed from known object geometry and planarity assumptions. 

\subsection{Implementation Details}
\label{sec:impl}

For numerical stability it is advised to avoid performing the direct matrix inversion of the covariance block matrix in Equation~\ref{eq:posteriorpredictive}\cite{rasmussen2006gaussian}. A robust alternative is to pre-factor the positive-definite covariance matrix using Cholesky decomposition $\Sigma(\mathbf{X}, \mathbf{X})^{-1} = \mathbf{L}\mathbf{L}^T$, and then use $\mathbf{L}$ and $\mathbf{L}^{-1}$ for subsequent calculations.

For our parallel GPU implementation we leverage the TensorFlow\cite{tensorflow2015-whitepaper} framework to obtain optimized tensor-products on GPUs. Computing the posterior predictive from Equation~\ref{eq:posteriorpredictive} decomposes nicely in matrix products. Additionally, the kernel formulation is already partly vectorized, see Equation~\ref{eq:covariancekernel}. Also note, that $\mathbf{L}$ and $\mathbf{L}^{-1}$ can be pre-computed on the CPU for a given training data set. The GPU then reads these values as constant matrices into GPU memory. 

We experimented with different versions of computing the kernel function. We found that the type of functions used had a heavy impact on execution speed. The best performance improvement compared to running on CPU was found to be one that makes explicit use of the following equivalence
\begin{equation}
	\begin{split}
		\lVert \mathbf{x_i} - \mathbf{x_j} \rVert^2 & = (\mathbf{x_i}-\mathbf{x_j})^T(\mathbf{x_i} -\mathbf{x_j}) \\
		 & = \mathbf{x_i}^T\mathbf{x_i} - 2\mathbf{x_i}^T\mathbf{x_j} + \mathbf{x_j}^T\mathbf{x_j}.
	\end{split}
\end{equation}

Additionally preloading data onto the GPU, while computing correction for previous batches, helps significantly in reducing GPU stalling due to I/O waits. For details consult our source code. 

\section{RESULTS}
In order to validate our findings, we captured depth data according to the procedure described in Section~\ref{sec:dataset}. The sensor has undergone a standard intrinsic and extrinsic calibration before capturing.

For the remainder of this work we refer to the device provided pixel-wise mean depth map at axis position $p$, temperature $t$ by $\Depth{ir}{p}{t}$. The mean depth map is computed by averaging over multiple captures (50 in our case) per axis position and temperature. Similarly, we refer to the artificial generated mean depth map from color images by $\Depth{rgb}{p}{t}$. Note that we assume $\Depth{ir}{p}{t}$ to be pre-aligned with the RGB device by means of methods presented in Section~\ref{sec:densedepthmaps}.

\begin{figure} [htp]
	\centering
	\subfloat[]{\includegraphics[width=0.5\textwidth]{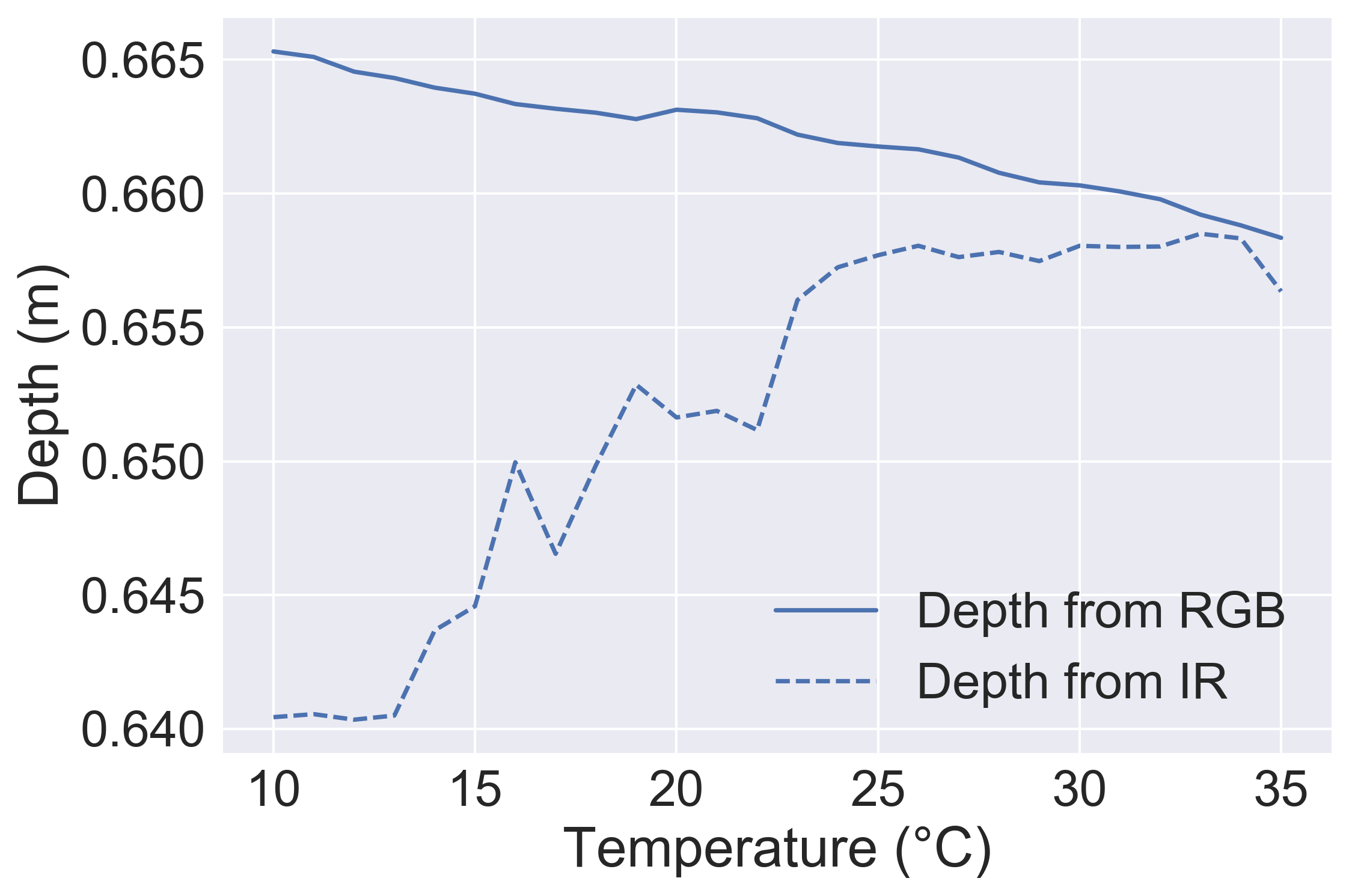}}
	\hfill
	\subfloat[]{\includegraphics[width=0.5\textwidth]{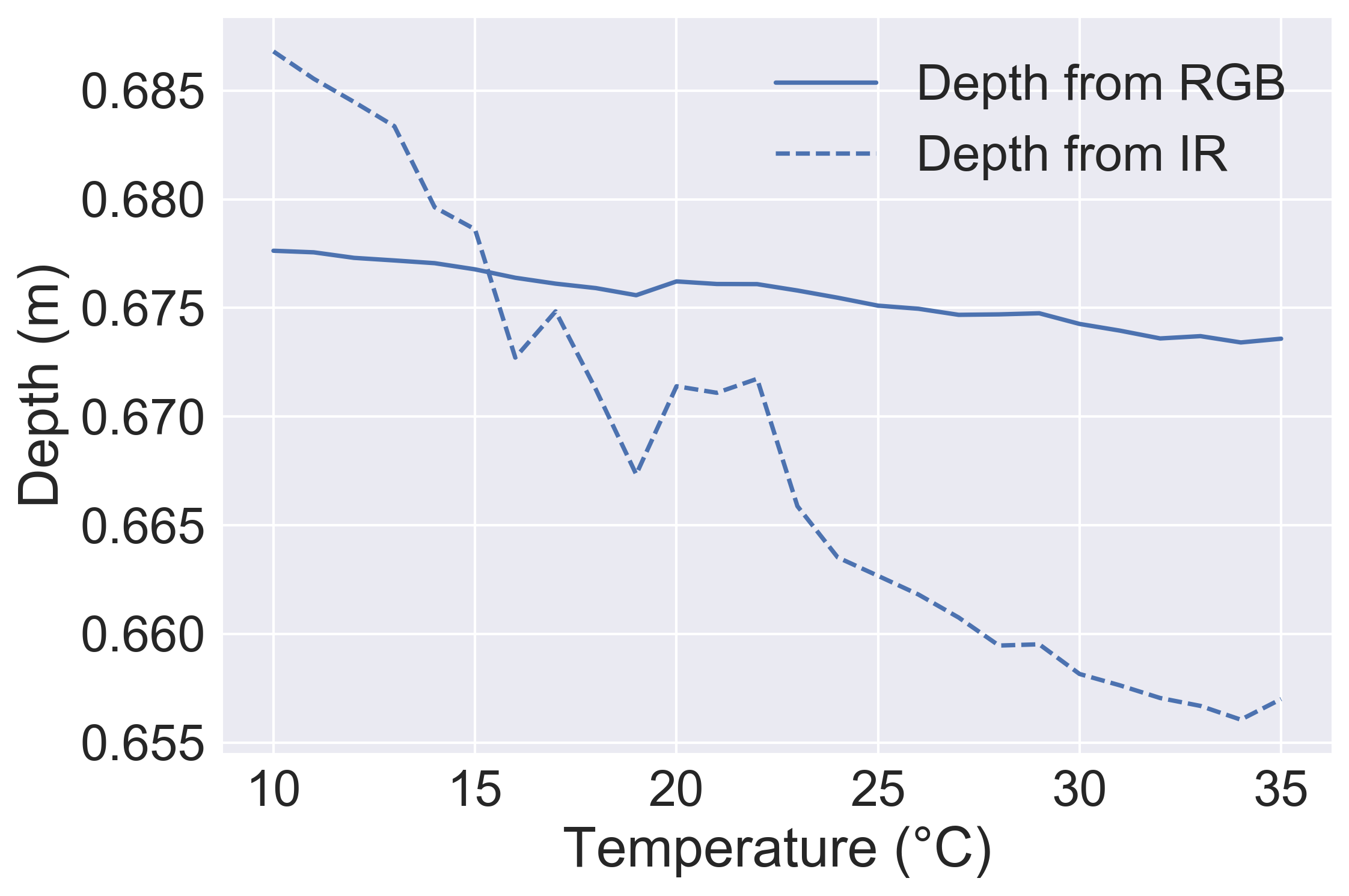}}
	\caption {
		\label{fig:tempinfluence} 
		Effects of varying temperature on depth estimation at a fixed distance over a small window region. (a) Unsteady depth estimates in a small window region located in the top-left image corner. (b) Unsteady depth estimates in a small window region located in the bottom-right image corner. One can see the high impact of temperature to the measured distance.
	}
\end{figure}

Figure \ref{fig:tempinfluence} illustrates the thermal influences on depth readings for a fixed position $\hat{p}$. The two plots are made from two small opposite image regions (top-left, bottom-right). One can clearly see the depth drift associated with $\Depth{ir}{\hat{p}}{t}$ is up to 4-5 times larger than the depth drift of $\Depth{rgb}{\hat{p}}{t}$. While the depth from RGB drift appears linear with increasing temperature, the depth readings from $\Depth{ir}{\hat{p}}{t}$ behave highly nonlinear. Additionally, the curves tend to flatten as we decrease the distance $p$ to the calibration target and escalate as we move further away from it.

For Gaussian Process Regression training, we sample input feature vectors $\mathbf{X}$ from a regularly spaced Cartesian 3D grid in meters, every three degrees Celsius. The corresponding target values $\mathbf{y}$ are then computed from the depth difference map given by 
\begin{equation}
	\label{eq:depthdiffmap}
	\Depth{delta}{p}{t} = \Depth{rgb}{p}{t_{\mathrm{min}}} - \Depth{rgb}{p}{t}. 
\end{equation}

That is, we always calibrate with respect to depth from color at the lowest possible temperature level. In total our training data consists of $\mathbf{X} \in \mathbb{R}^{5000\times4}$ elements and $\mathbf{y} \in \mathbb{R}^{5000\times1}$ elements. We then auto-tune the kernel hyper-parameters $\begin{bmatrix}\mathbf{W},\sigma_y,\sigma_s\end{bmatrix}$ using gradient descent of the negative log-marginal-likelihood\cite{murphy2012machine}, yielding the following optimal parameters for our dataset: $diag(\mathbf{W})$ is given by $\begin{bmatrix}2.0,0.04,1.29,0.002\end{bmatrix}$, $\sigma_y=0.044$ and $\sigma_s=0.031$ for our dataset. As expected, our formulation of $\Depth{delta}{p}{t}$ results in high importance along the temperature domain, making samples far apart in the temperature domain still influential on the regressed depth correction. 

\begin{figure} [htp]
	\centering
	\subfloat[{t=\unit[12]{\degree C}, p=\unit[60]{cm}}] {\includegraphics[width=1.0\textwidth]{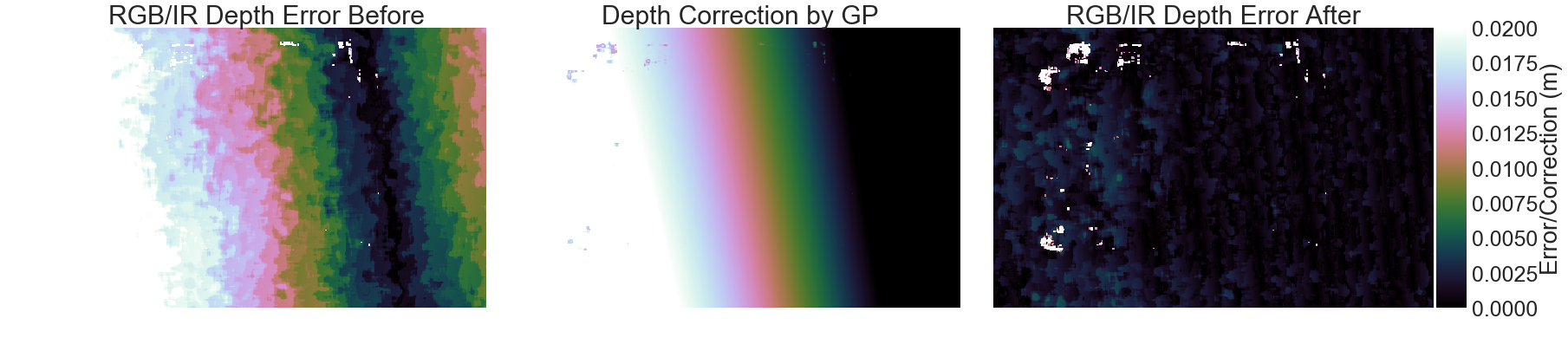}} \hfill \\
	\subfloat[{t=\unit[17]{\degree C}, p=\unit[40]{cm}}]{\includegraphics[width=1.0\textwidth]{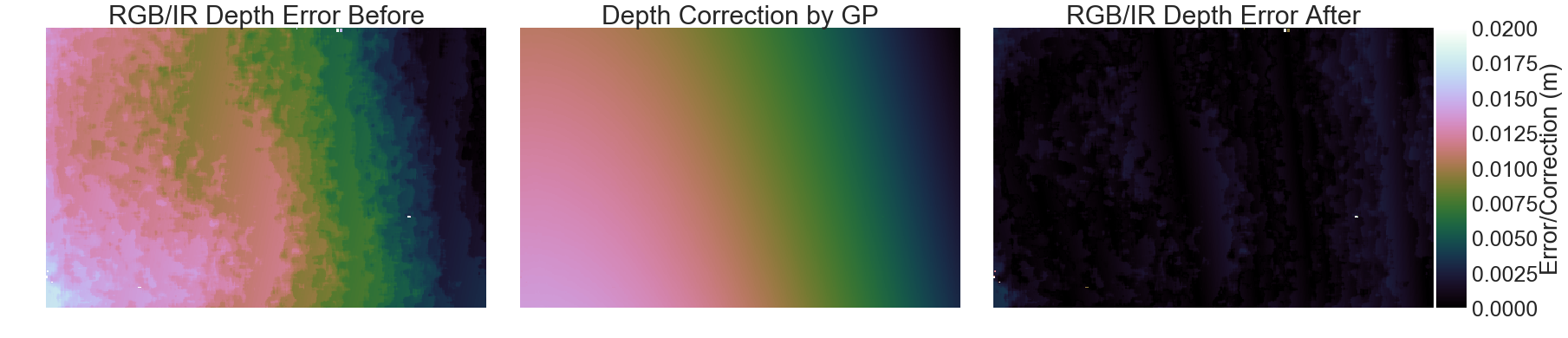}} \hfill \\
	\subfloat[{t=\unit[22]{\degree C}, p=\unit[40]{cm}}]{\includegraphics[width=1.0\textwidth]{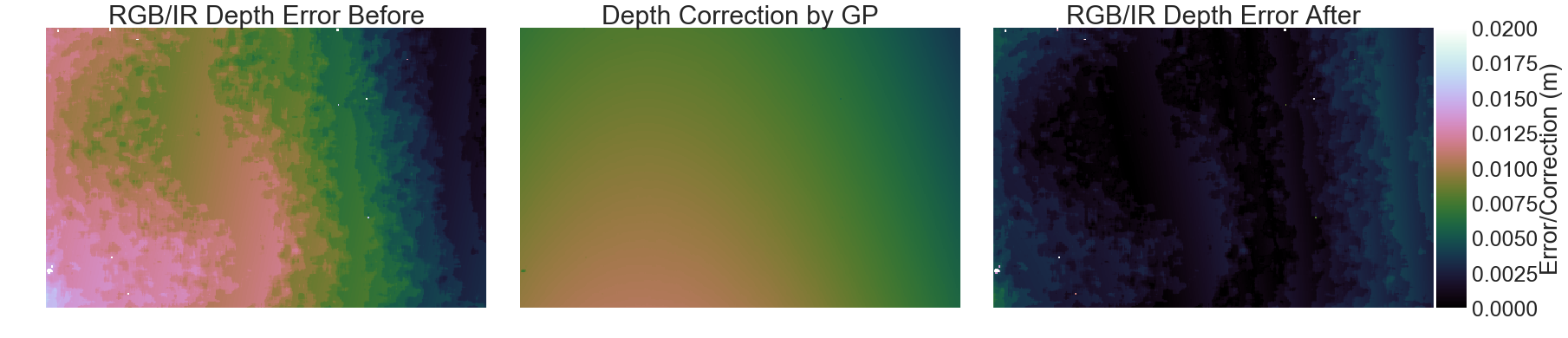}} \hfill \\
	\subfloat[{t=\unit[25]{\degree C}, p=\unit[90]{cm}}]{\includegraphics[width=1.0\textwidth]{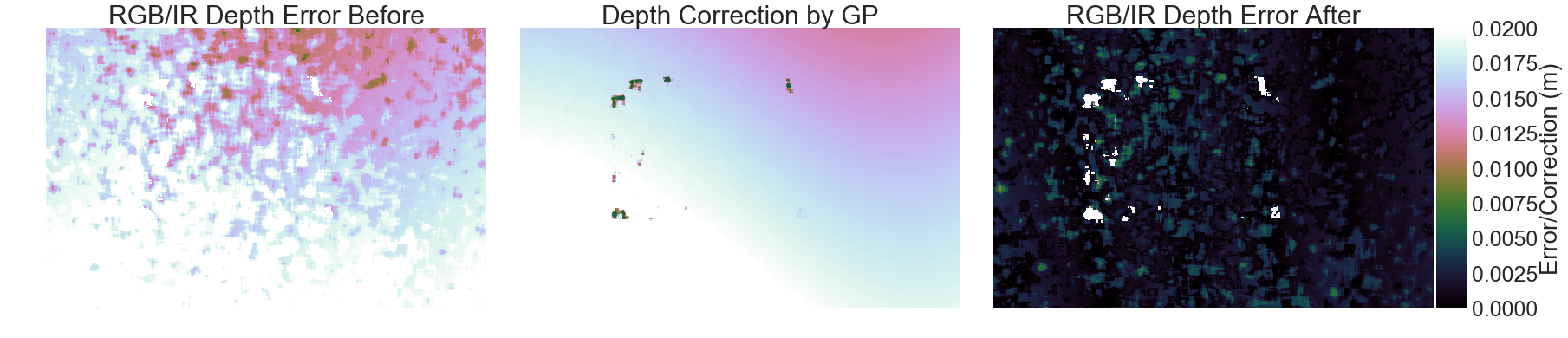}} \hfill \\
	\subfloat[{t=\unit[35]{\degree C}, p=\unit[60]{cm}}]{\includegraphics[width=1.0\textwidth]{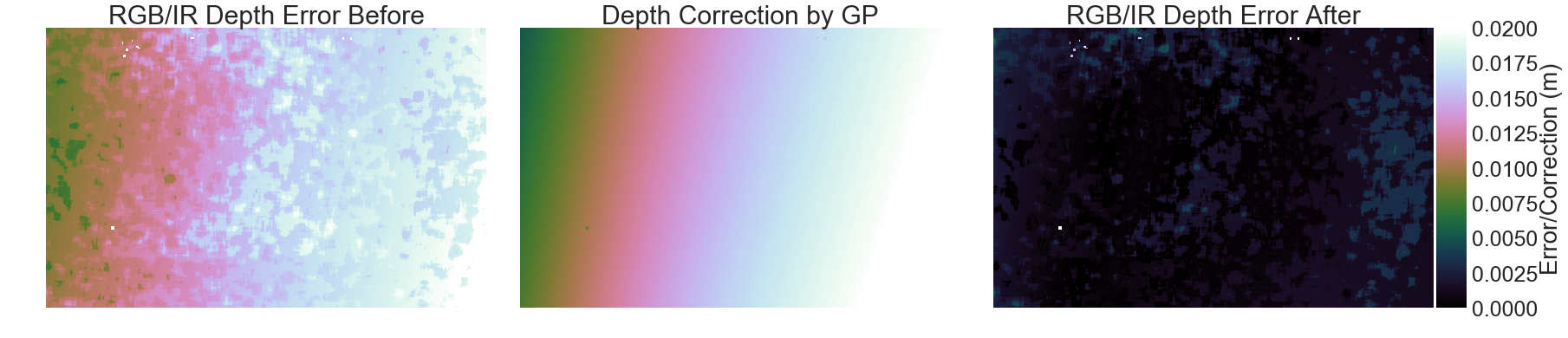}} \hfill \\
	\caption {
		\label{fig:correction} 
		Depth correction by Gaussian Process Regression. Subfigures (a-e) refer to before/after correction effects for varying positions and temperatures. The white-speckles in after-images are due to missing depth sensor readings for which no correction can be computed. The error is significantly reduced independent from chosen position and temperature.
	}
\end{figure}

Applying the proposed Gaussian Process Regression reduces the error significantly. The plots in Figure \ref{fig:correction} show the pixel-wise error $\Depth{rgb}{p}{t_{\mathrm{min}}} - \Depth{rgb}{p}{t}$ before and after application of the regression for a selected set of positions and temperatures. For areas of missing depth data no predictions can be made. Those pixels appear as white speckles in the illustrations. A numerical evaluation spanning the entire distance and temperature range is provided in Table~\ref{tab:rmse}. One can see that the correction yields an improvement by one order of magnitude.

\begin{table}[htp]	
	\centering
	\caption{\label{tab:rmse} $\Depth{rgb}{p}{t_{\mathrm{min}}} - \Depth{rgb}{p}{t}$ RMSE before and after correction in Cartesian space across all temperature and distance captures.  The correction yields an improvement by one order of magnitude.}
	
	\begin{tabular}{l|c|c|c|}
	\cline{2-4}
															& \multicolumn{1}{l|}{\textbf{$x$ (mm)}} & \multicolumn{1}{l|}{\textbf{$y$ (mm)}} & \multicolumn{1}{l|}{\textbf{$z$ (mm)}} \\ \hline
	\multicolumn{1}{|l|}{\textbf{RMSE before correction}} & 5.7                                    & 3.7                                    & 16.0                                   \\ \hline
	\multicolumn{1}{|l|}{\textbf{RMSE after correction}}   & 0.7                                    & 0.5                                    & 2.2                                    \\ \hline
	\end{tabular}
\end{table}

Table \ref{tab:performance} compares execution times for CPU and GPU variants on dense depth maps consisting of over 300,000 points as described in Section~\ref{sec:impl}. The GPU variants clearly outperform an optmized CPU implementation. When possible, one should prefetch data to reduce wait times and enable real-time performance.

\begin{table}[htp]
	\centering
	\caption{\label{tab:performance} Execution times for correction per depth map frame of size 640 times 480. The GPU optimized version outperforms the other variants.} 
	\begin{tabular}{l|c|c|}
	\cline{2-3}
												& \multicolumn{1}{l|}{\textbf{Time (s)}} & \multicolumn{1}{l|}{\textbf{FPS (1/s)}} \\ \hline
	\multicolumn{1}{|l|}{\textbf{CPU}}          & 20                                     & 0.05                                    \\ \hline
	\multicolumn{1}{|l|}{\textbf{GPU na\"{\i}ve}}          & 0.4                                    & 2.5                                     \\ \hline
	\multicolumn{1}{|l|}{\textbf{GPU optimized}} & 0.14                                   & 7.1                                     \\ \hline
	\end{tabular}
\end{table}

To show generality of our findings we captured and evaluated data from two equivalent sensors. However, the results differed only marginally.

\section{CONCLUSION AND FUTURE WORK}
In this paper we demonstrated the thermal influence on active structured-light RGB-D cameras based on Orbbec Astra Mini S sensors. A novel depth-delta calibration method, based on probabilistic regression in the spatial and thermal domains, was presented. We showed that Gaussian Progress Regression is well suited to correct depth errors under spatio-thermal changes. The comprehensive evaluation of the proposed approach on a novel dataset demonstrates a reduction of the root mean squared error by one order of magnitude. A further comparison of CPU and GPU implementations proves the real-time performance capabilities of our approach. The dataset and source code was made publicly available.

Possible future work concerns the evaluation of sensors such as time-of-flight based models. Regarding performance, a scale-space approach could enable real-time performance directly on embedded systems. Additionally, fixed point or half-precision variants could be examined to further relax computational runtime requirements. 

\section{ACKNOWLEDGEMENTS}
This research is funded by the projects Lern4MRK (Austrian Ministry for Transport, Innovation and Technology), and AssistMe (FFG, 848653), as well as the European Union in cooperation with the State of Upper Austria within the project Investition in Wachstum und Besch\"aftigung (IWB). The authors want to thank Mr. Gerhard Ebenhofer for fruitful discussions and help during the data acquisition.

\bibliography{main} 
\bibliographystyle{spiebib} 

\end{document}